\documentclass[letter,onecolumn,11pt]{article}

\usepackage[letterpaper,margin=1in]{geometry}

\usepackage{microtype}

\usepackage{times}
\usepackage{helvet}
\usepackage{courier}
\usepackage{mathtools}
\usepackage{graphicx}
\usepackage{multirow}
\usepackage{url}
\graphicspath{ {images/} }
\usepackage{graphicx} 
\usepackage{subfigure} 
\usepackage{lipsum}
\usepackage{caption}
\usepackage{mwe}
\usepackage[noend]{algpseudocode}
\usepackage{algorithm}
\usepackage{enumitem}
\usepackage{color}
\usepackage{amsmath}
\usepackage{amssymb}
\usepackage{xspace}
\usepackage{amsfonts}
\usepackage{booktabs}
\usepackage{tabularx}
\usepackage{arydshln} 
\usepackage[T1]{fontenc}
\usepackage[utf8]{inputenc}
\usepackage{graphicx}

\usepackage[dvipsnames]{xcolor}

\usepackage{tikz}
\usepackage{pgfplots}

\usepackage{array,tabularx}
\newcolumntype{R}[1]{>{\raggedleft\arraybackslash}p{#1}}
\newcolumntype{L}[1]{>{\raggedright\arraybackslash}p{#1}}
\newcolumntype{C}[1]{>{\centering\arraybackslash}p{#1}}
\newcolumntype{A}{>{\raggedright\arraybackslash}X}
\newcolumntype{B}[1]{>{\raggedright\arraybackslash}b{#1}}
\usepackage[flushleft]{threeparttable}

\usepackage{calc}

\usepackage{listings}
\lstdefinestyle{MySQL}{
  breaklines=true,
  language=SQL,
  frame=ltrb,
  framesep=5pt,
  basicstyle=\footnotesize,
  keywordstyle=\ttfamily\color{OliveGreen},
  identifierstyle=\ttfamily\color{CadetBlue}\bfseries,
  commentstyle=\color{Brown},
  stringstyle=\ttfamily,
  showstringspaces=false
}

\colorlet{punct}{red!60!black}
\definecolor{background}{HTML}{EEEEEE}
\definecolor{delim}{RGB}{20,105,176}
\colorlet{numb}{magenta!60!black}

\lstdefinelanguage{JavaScript}{
  keywords={typeof, new, true, false, catch, function, return, null, catch, switch, var, if, in, while, do, else, case, break},
  keywordstyle=\color{blue}\bfseries,
  ndkeywords={class, export, boolean, throw, implements, import, this},
  ndkeywordstyle=\color{darkgray}\bfseries,
  identifierstyle=\color{black},
  sensitive=false,
  comment=[l]{//},
  morecomment=[s]{/*}{*/},
  commentstyle=\color{purple}\ttfamily,
  stringstyle=\color{red}\ttfamily,
  morestring=[b]',
  morestring=[b]"
}
\usepackage{tikz}
\usepackage{pgfplots}
\usepackage[export]{adjustbox}
\usepackage{framed}
\setlist[itemize]{leftmargin=*,itemsep=2ex}
\usepackage{titlesec}

\newcommand{\sme}{\textit{subject matter expert}\xspace}

\newcommand{\dis}{\textit{discovery}\xspace}

\newcommand{\mlt}{\textbf{ML 2.0}\xspace}
\newcommand{\mlo}{\textbf{ML 1.0}\xspace}
\newcommand{\dev}{developer \xspace}

\newcommand{\lf}{\textit{labeling} \xspace}
\newcommand{\fm}{$\mathtt{feature\_matrix}$ \xspace}
\newcommand{\pred}{$\mathtt{predictions}$ \xspace}

\newcommand{\api}{$\mathtt{API}$\xspace}
\newcommand{\ES}{\textit{Entityset}\xspace}
\newcommand{\es}{Entityset\xspace}
\newcommand{\esm}{$\mathtt{Entityset}$\xspace}
\newcommand{\atm}{$\mathtt{ATM}$\xspace}
\newcommand{\ft}{$\mathtt{Featuretools}$\xspace}
\newcommand{\cof}{$\mathtt{cutoff\_time}$\xspace}
\newcommand{\fl}{$\mathtt{feature\_list}$\xspace}
\newcommand{\ind}{$\mathtt{index}$\xspace}
\newcommand{\tind}{$\mathtt{time index}$\xspace}

\newcommand{\Sme}{Subject matter expert\xspace}
\newcommand{\modeldesc}{$\mathtt{model\_provenance.json}$\xspace}
\newcommand{\metajson}{$\mathtt{metadata.json}$\xspace}
\PassOptionsToPackage{hyphens}{url}
\usepackage[bookmarks=false,colorlinks=true,linkcolor={blue},citecolor={blue},urlcolor={blue},breaklinks=true]{hyperref}
\usepackage[numbers]{natbib}

\usepackage{tcolorbox}

\tcbset{colback=white,colframe=gray!75!black}

\newcounter{mybox}
\newenvironment{mybox}[1][\unskip]{%
\begin{tcolorbox}
\stepcounter{mybox}
\bgroup\sffamily(\arabic{mybox})\hspace{1em} \textbf{\Large #1}\egroup
\tcblower
}
{\end{tcolorbox}}

\usepackage{glossaries}

\makeglossaries

\newglossaryentry{machine learning}
{
name={machine learning},
description={Machine learning is a field of computer science that gives computers the ability to learn without being explicitly programmed}
}

\newglossaryentry{warehouse}
{
name={warehouse},
description={this is an example}
}

\newglossaryentry{deep feature synthesis}
{
name={deep feature synthesis},
description={this is an example}
}

\newglossaryentry{label times}
{
name={label times},
description={this is an example}
}

\usepackage{endnotes}

\makeatletter
\def\@maketitle{%
\begin{center}
{\LARGE
Machine learning 2.0\\ Engineering data driven AI products
}
\vskip0.3in
\begin{minipage}{0.3\linewidth}
\centering
{\Large Max Kanter}\\\vskip3pt
{\large Feature Labs, Inc. \par Boston, MA 02116}\\\vskip3pt
\texttt{\small max.kanter@featurelabs.com}
\end{minipage}
\hfill
\begin{minipage}{0.3\linewidth}
\centering
{\Large Benjamin Schreck}\\\vskip3pt
{\large Feature Labs, Inc. \par Boston, MA 02116}\\\vskip3pt
\texttt{\small ben.schreck@featurelabs.com}
\end{minipage}
\hfill
\begin{minipage}{0.3\linewidth}
\centering
{\Large Kalyan Veeramachaneni}\\\vskip3pt
{\large MIT LIDS, \par Cambridge, MA 02139}\\\vskip3pt
\texttt{\small kalyanv@mit.edu}
\end{minipage}
\vskip0.3in
\textcolor{gray}{\small v0.1.0}\\
\textcolor{gray}{\small Date: 2018-03-06}
\end{center}
\vskip0.3in
}%
\makeatother

\begin{document}


\maketitle
\date{}

\begin{abstract}
\textbf{\mlt}: In this paper, we propose a paradigm shift from the current practice of creating machine learning models -- which requires months-long discovery, exploration and ``feasibility report'' generation, followed by re-engineering for deployment -- in favor of a rapid, 8-week process of  development, understanding, validation and deployment that can executed by developers or \sme{s} (non-ML experts) using reusable \api{s}. This accomplishes what we call a ``minimum viable data-driven model,'' delivering a ready-to-use machine learning model for problems that haven't been solved before using machine learning.  We provide provisions for the refinement and adaptation of the ``model," with strict enforcement and adherence to both the scaffolding/abstractions and the process. We imagine that this will bring forth a second phase in machine learning, in which \textit{discovery} is subsumed by more targeted goals of delivery and impact.
\end{abstract}

\section{Introduction}
Attempts to embed machine learning-based predictive models into products and services in order to make them smarter, faster, cheaper, and more personalized will dominate the technology industry for the foreseeable future. Currently, these applications aid industries ranging from financial services systems, which often employ simple fraud detection models, to patient care management systems in intensive care units, which employ more complex models that predict events. If the many research articles, news articles, blogs, and data science competitions based on new data-driven \textit{discoveries} are to be believed, future applications will be fueled by applying machine learning to numerous data stores — medical (see \cite{avati2017improving}, \cite{shi2017towards}), financial and others. But it is arguable how many of the predictive models described in these announcements have actually been deployed — or have been effective, serving their intended purposes of saving costs, increasing revenue and/or enabling better experiences \cite{wagstaff2012machine}.  \footnote{The authors of \cite{sculley2014machine} offer one perspective on the challenges of deploying and maintaining a machine learning model, and others in the tech industry have highlighted these challenges as well. The core problem, however, goes deeper than deployment challenges.} 

To explicate this, we highlight a few important observations about how machine learning and AI systems are currently built, delivered and deployed \footnote{These observations have been gathered from our experiences creating, evaluating, validating, publishing and delivering machine learning models using data from \textsc{BBVA}, \textsc{Accenture}, \textsc{Kohls}, \textsc{Nielsen}, \textsc{Monsanto}, \textsc{Jaguar and Land Rover}, \textsc{edX}, MIMIC, \textsc{GE}, \textsc{Dell}, and others. Additionally, we have entered our automation tools into numerous publicly-held data science competitions held by \textsc{Kaggle} and similar websites. We are also a part of the MIT team involved with the DARPA D3M initiative, and development of the automation system Featuretools is funded under DARPA. The findings and opinions expressed here are ours alone, and do not represent any of our clients, funders, or collaborators.}. In most cases, the development of these models involves the following criteria: 1. It relies on making an initial \textit{discovery} from the data; 2. It uses a historically defined and deeply entrenched workflow; and 3. It struggles to find the functional interplay between a robust software engineering practice and the complex landscape of mathematical concepts required to build the models 
We call this current state of machine learning \textbf{ML 1.0}, defined by its focus on \textit{discovery}. 
\footnote{Throughout this paper, our core focus is on data that is temporal, multi-entity, multi-table, and relational (and/or transactional). In most cases, we are attempting to predict using a machine learning model, and in some cases we are predicting ahead of time.}


\textbf{The ``discovery first'' paradigm}:  Most of the products within which we are currently trying to embed intelligence have already been collecting data as part of their normal, day-to-day functioning. Machine learning or predictive modeling is usually an afterthought. Machine learning models are attempting to predict a future outcome -- one that is legible to the humans using it -- and it is not always certain whether the data at hand will be able to provide such conclusions.



This makes machine learning unlike other software engineering projects that involve developing or adding a new feature to a product. In those cases, the end outcome is deterministic and visible. Designers, architects, and managers can make a plan, establish a workflow, release the feature, and manage it. 

In contrast, a new machine learning project usually starts with a \textit{discovery} phase, in which attempts are made to answer the quintessential question: \textit{``Is this predictable using our data?"}.  If the answer is yes, further questions emerge: With how much accuracy? What variables mattered? Which one of the numerous modeling possibilities works best? If the data happens to be temporal, which one of the numerous time series models (with latent states) can model it best? What explainable patterns did we notice in the data, and what models surfaced those patterns better? And so on, in an endless list. 


It is not uncommon to see dozens of research papers focused on the same prediction problem, each providing slightly different answers to any of the questions above. A recently established prediction problem -- ``predicting dropout in Massive Open Online Courses, to enable interventions'' -- resulted in at least 100 research papers, all written in a span of 4 years\footnote{These papers are published in premier AI venues -- NIPS, IJCAI, KDD, and AAAI -- and one of our first predictive modeling projects, for weblog data in 2013-14, focused on this very problem}, and a competition at a Tier 1 data mining conference, KDD Cup 2015 \cite{sinha2014your, ramesh2013modeling, he2015identifying, halawa2014dropout}.


The expectation is that once this question has been answered, a continuously working predictive model can be developed, integrated with the product, and put to use rather quickly. This expectation might not be so unrealistic if it weren't for the deeply entrenched workflow generally used for \textit{discovery}, as we describe below.

\begin{figure*}
\centering
\includegraphics[width=0.86\linewidth]{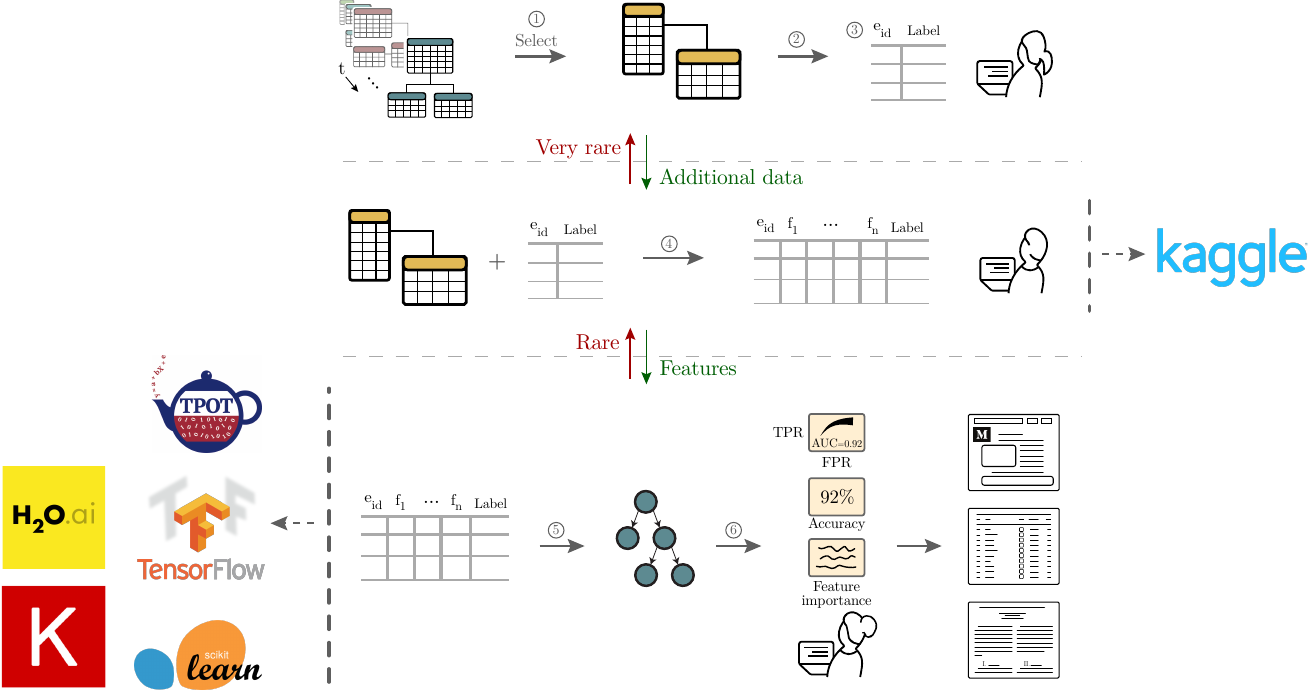}
\caption{ML 1.0 workflow. This traditional workflow is split into three disjoint parts, each of which contains a subset of steps, which are often executed by different people. Intermediate data representations generated at the end of each part are stored and passed along to the next people in the chain. At the top, a relevant subset of the data is extracted from the data warehouse, the prediction problem is decided, past training examples are identified within the data, and the data is prepared so as to enable model building whilst avoiding issues like ``label leakage." This step is usually executed in collaboration with people that manage/collect the data (as well as the product that creates it), domain/business experts who help define the prediction problem, and data scientists who use the data to build the predictive models. The middle part of the workflow is feature engineering, where, for each training example, a set of features is extracted by the data scientists. If statistics and machine learning experts are involved, this part could be executed by software engineers or data engineers or the data could be released as a public competition on \textsc{Kaggle}. The last part involves of building, validating, and analyzing the predictive model. This part is executed by someone familiar with machine learning. A number of automated tools and open-source libraries are available to do help with this part, including a few shown on the left. The goal of this process has mostly been \textit{discovery}, ultimately resulting in a \textit{research paper}, a \textit{competition leader board}, or a \textit{blog post}. In these cases, putting the machine learning model to use is an afterthought. This traditional workflow has a couple of problems: decisions made during previous parts are rarely revisited; the prediction problem is decided early-on, without an exploration of how small changes in it can garner significant benefits; and end-to-end software cannot be compiled in one place so that it can be repeated with new data, or lead to the creation of deployable model.  } \label{fig:current_workflow}
\end{figure*}

\noindent \textbf{A ``deeply entrenched workflow''}: Developing and evaluating machine learning models for industrial-scale problems predates the emergence of internet-scale data. Our observations of machine learning in practice over the past decade, as well as our own experiences developing predictive models, have enabled us to delineate a number of codified steps that make up these projects. 
These steps are shown in Table ~\ref{table-steps}. Over the years, surprisingly little has changed about how researchers approach a data store. For example, you can easily compare the structure of these three papers, one written in 1994 (\cite{ghosh1994credit}) and two written in 2017 (\cite{avati2017improving, shi2017towards}).
\noindent \textit{So what is the problem?}: Although the generalized, codified steps shown in Table ~\ref{table-steps} make up a good working template, the workflow in which they are executed contains problems that are now entrenched. In Figure~\ref{fig:current_workflow}, we depict how these steps are currently executed as if they are three disjoint parts. Recent tools and collaborative systems tend to apply to just one of these parts, accelerating discovery only when the data is ready to be used in that part. We present our detailed commentary on the current state of applied machine learning in Section~\ref{commentary}. 


\begin{table*}[!t]
\begin{tabularx}{\linewidth}{@{}p{2.2cm}X@{}}
\includegraphics[width=2cm,valign=t]{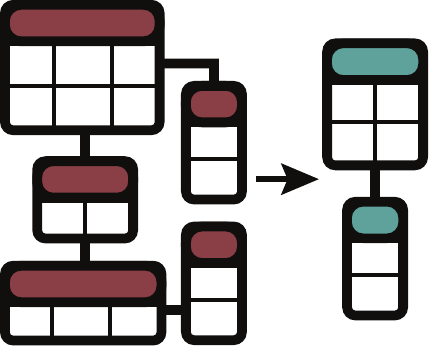} & \textbf{1.\ Extract relevant data subset}\par
Data warehouses (usually centrally organized) contain several data elements that may not be related to the predictive problem at hand, and may cover a large time period. In this step, a subset of tables and fields is determined, and a specific time period is selected. Based on these two choices, a number of filters are applied, and the resulting data is passed on to Step~2. 
\\[2em]
\includegraphics[width=2cm,valign=t]{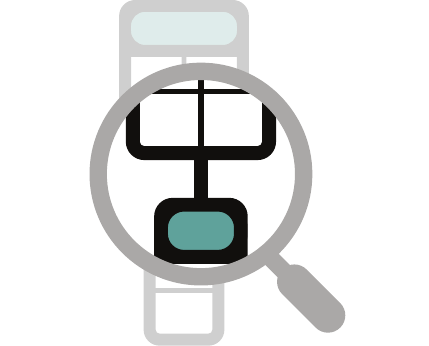} & \textbf{2.\ Formulate problem and assemble training examples}\par
Developing a model that can predict future events requires finding past examples to learn from. Typical predictive model development thus involves first defining the outcome in question, and then finding past occurrences of that outcome that could be used to learn a model. \\[2em]
\includegraphics[width=2cm,valign=t]{Icon2.pdf} & \textbf{3.\ Prepare data}\par
To train a model, we use data retrospectively to emulate the prediction scenario -- that is, we use data prior to the occurrence of the outcome to learn a model, and to evaluate its ability to predict the outcome. This requires careful annotation regarding which data elements can be used for modeling. In this step, most of these annotations are added, and the data is filtered to create a dataset ready for machine learning. \\[2em]
\includegraphics[width=2cm,valign=t]{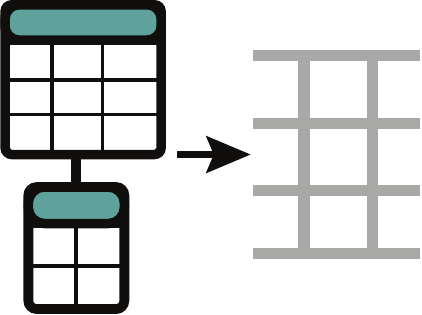} & \textbf{4.\ Engineer features}\par
For each training example, given the usable data, one computes features (aka variables) and creates a machine learning-ready matrix. Each column in the matrix is a feature, the last column is the label, and each row is a different training example. \\[2em]
\includegraphics[width=2cm,valign=t]{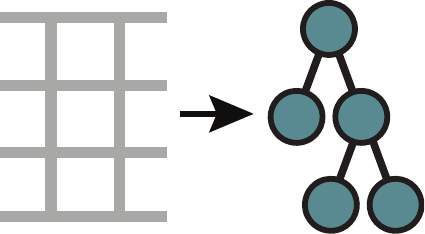} & \textbf{5.\ Learn a model}\par
Given the feature matrix, in this step a model -- either classifier (if the outcome is categorical) or regressor (if the outcome is continuous) -- is learned. Numerous models are evaluated and selected based on a user-specified evaluation metric. \\[2em]
\includegraphics[width=2cm,valign=t]{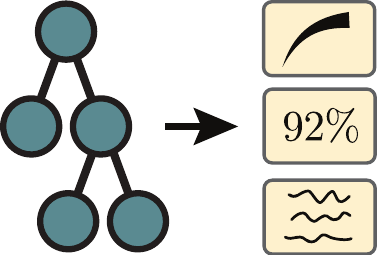} & \textbf{6.\ Evaluate the model and report}\par
Once a modeling approach has been selected and trained, it is evaluated using the training data. A number of metrics, including precision and recall, are reported, and the impact of variables on predictive accuracy is evaluated.
\end{tabularx}
\caption{Steps to making a \textit{discovery} using machine learning from a data warehouse. This process does not define a robust software engineering practice, nor does it account for model testing or deployment. The end result of the process is to generate a report or research paper. The typical workflow breaks the process down into silos, presented in Figure~\ref{fig:current_workflow}.  }\label{table-steps}
\end{table*}

\begin{figure*}[!htb]
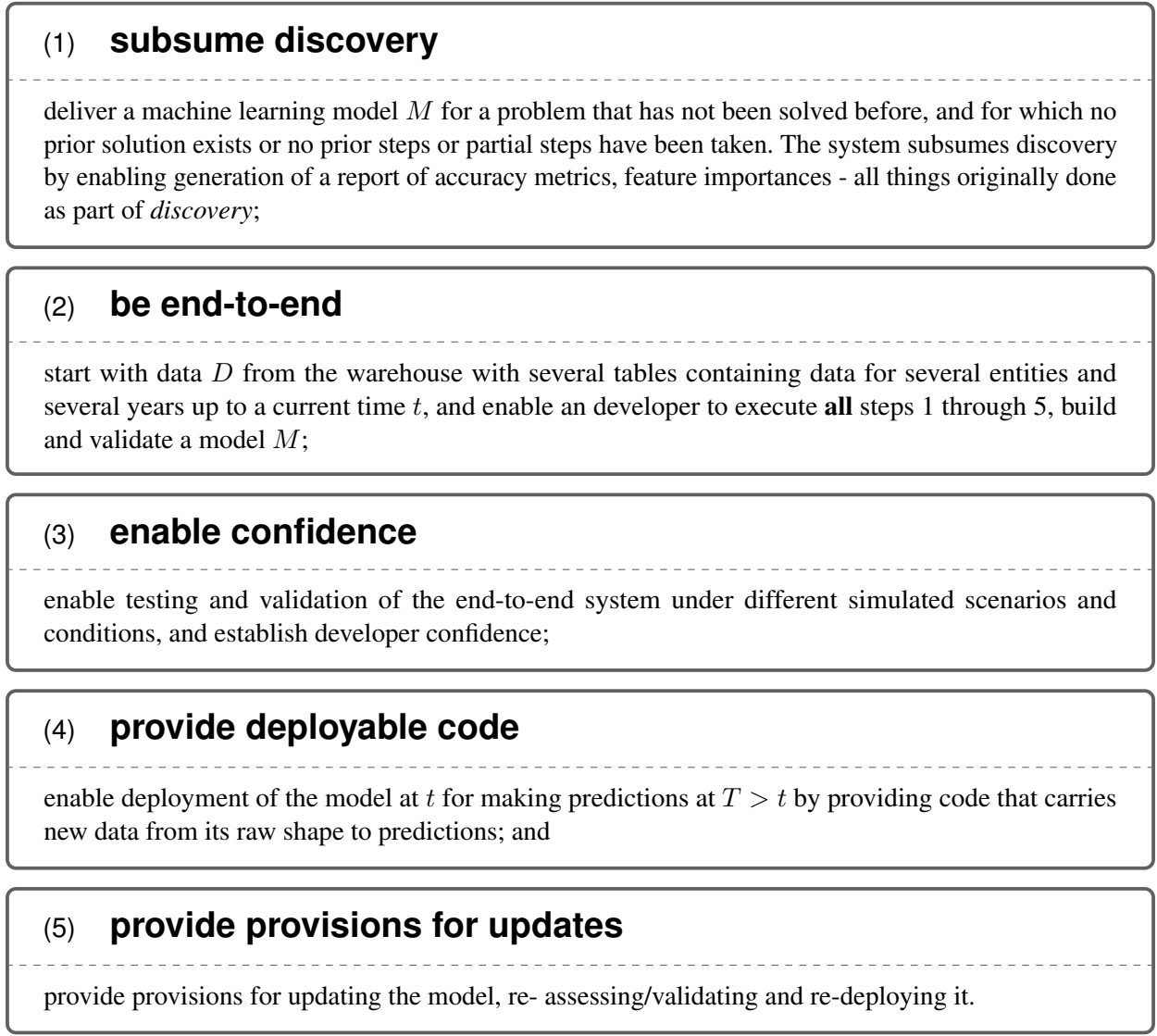


\begin{mybox}[subsume discovery]
deliver a machine learning model $M$ for a problem that has not been solved before, and for which no prior solution exists or no prior steps or partial steps have been taken. The system subsumes discovery by enabling generation of a report of accuracy metrics, feature importances - all things originally done as part of \textit{discovery};
\end{mybox}

\begin{mybox}[be end-to-end]
start with data $D$ from the warehouse with several tables containing data for several entities and several years up to a current time $t$, and enable an developer to execute \textbf{all} steps 1 through 5, build and validate a model $M$;
\end{mybox}

\begin{mybox}[enable confidence]
enable testing and validation of the end-to-end system under different simulated scenarios and conditions, and establish developer confidence;
\end{mybox}

\begin{mybox}[provide deployable code]
enable deployment of the model at $t$ for making predictions at $T>t$ by providing code that carries new data from its raw shape to predictions; and
\end{mybox}

\begin{mybox}[provide provisions for updates]
provide provisions for updating the model, re- assessing/validating and re-deploying it.
\end{mybox}
\caption{Goals and requirements for \mlt}
\label{goals}
\end{figure*}

\noindent \textbf{ML 2.0: Delivery and impact}: In this paper, we propose a paradigm shift, turning from the current practice of creating machine learning models -- which involves a months-long process of discovery, exploration, ``feasibility report'' generation, and re-engineering for deployment -- in favor of a rapid, 8-week-long process of development, understanding, validation and deployment that can executed by developers or \sme{s} (non-ML experts) using reusable \api{s}. This accomplishes what we call a ``minimum viable data-driven model,'' delivering a ready-to-use machine learning model for problems that haven't been solved before using machine learning. 

We posit that any system claiming to be \mlt should deliver on the key goals and requirements listed in Figure~\ref{goals}. Each of these steps is important for different reasons: (1) ensures that we are bringing new intelligence services to life, (2) requires that the system considers the entire process and not just one step, and (3), (4) and (5) guarantee the model's deployment and use.

\noindent All of the above steps should:
\begin{itemize}
    \item[--] require a minimal amount of coding (possibly with simple \api calls);
    \item[--] use the same abstractions/software; 
    \item[--] quickly deliver the first version of the model;
    \item[--] require a minimal amount of manpower and no ``machine learning expertise.'' 
\end{itemize}

This will ensure a reduction in design complexity and speed up the democratization of AI.  We imagine that this will bring forth a second phase in artificial intelligence, in which the elevation of \textit{discovery} is subsumed by more targeted goals of delivery and impact.

\section{The path that led to ML 2.0}
Over the last decade, the demand for predictive models has grown at an increasing rate. As the data scientists who build these models, we have found that our biggest challenge typically isn't creating accurate solutions for incoming prediction problems, but rather the time it takes as to build an end-to-end model, which makes it difficult to match the supply of expertise to the incoming demand. We have also seen, disappointingly, that even when we build models, they are often not deployed. ML 2.0 is the result of a series of efforts to address these problems by automating those aspects of creating and deploying machine learning solutions that have, in the past, involved time- and skill-intensive human effort.

In 2013, we focused our efforts on the problem of selecting a machine learning method and then tuning its hyperparameters. This eventually resulted in Auto-Tuned Models (ATM)\cite{swearingenatm}\footnote{https://github.com/HDI-Project/ATM}, which takes advantage of cloud-based computing to perform a high-throughput search over modeling options and find the best possible modeling technique for a particular problem\footnote{ATM was published and open-sourced in 2017 to aid in the release of ML 2.0, but was originally developed in 2014}. 

While ATM helped us quickly build more accurate models, we couldn't take advantage of it unless the data we were using took the form of a feature matrix. As most of the real-world use cases we worked on did not provide data in such a form, we decided to move our focus earlier in the data science process. This led us to explore automating feature engineering, or the process of using domain-specific knowledge to extract predictive patterns from raw datasets. Unlike most data scientists who work in a single domain, we worked with a wide range of industries. This gave us the unique opportunity to develop innovative solutions for each of the diverse problems we faced. Across industries including finance, education, and health care, we started to see commonalities in how features were constructed. These observations led to the creation of the Deep Feature Synthesis (DFS) algorithm for automated feature engineering \cite{kanter2015deep}. 

We put DFS to the test against human data scientists to see if it accomplished the goal of saving us significant time while preparing raw data. We used DFS to compete in 3 different worldwide data science competitions and found that we could build models in a fraction of the time it took human competitors, while achieving similar predictive accuracies. 

After this success, we took our automation algorithms to industry. Now that we could quickly build accurate models using just a raw dataset and a specific problem, a new automation question arose -- ``How do we figure out what problem to solve?'' This question was particularly pertinent to companies and teams new to machine learning, who frequently had a grasp of high-level business issues, but struggled to translate them into specific prediction problems.

We discovered that there was no existing process for systematically defining a prediction problem. In fact, by the time labels reached data scientists, they had typically been reduced to a list of true / false values, with no annotations regarding their origins. Even if a data scientist had access to the code that extracted the labels, it was implemented such that it could not be tweaked based on domain expert feedback or reused for different problems. To remedy this, we started to explore ``prediction engineering,'' or the process of defining prediction problems in a structured way. In a 2016 paper \cite{kanter2016label}, we laid out the Label Segment Featurize (LSF) abstraction and associated $\mathtt{traversal}$ algorithm to search for labels.

As we explored prediction engineering and defined the LSF abstraction, the concept of $\mathtt{time}$ began to play a huge role. Because they involve both simulating past predictive scenarios and carefully extracting training examples in order to prevent label leakage while training, predictive tasks are intricately tied to time. This inspired perhaps the most important requirement -- a way to annotate every data point with the timestamp at which it occurred. We then revisited each data-processing algorithm we had written for feature engineering, prediction engineering, or preprocessing and programmed it to accept only time periods where data was valid by introducing a simple yet powerful notion: \cof. When \cof is specified for a data processing block, any data that ``occurred'' beyond it cannot be used.

At the same time, it became clear that we needed metadata in order to perform automatic processing and maintain consistency throughout a modeling endeavor. We defined a relational representation for pandas dataframes, called \es, and started to define operations over it (described in Section~\ref{es}). To enable easy access to the metadata, we defined a \metajson \footnote{Our current version of this $\mathtt{json}$ is available at \url{https://github.com/HDI-Project/MetaData.json}}. We also started to define a more concrete way to organize the meta-information about the processing done on the data from a predictive modeling standpoint, which led to the development of \modeldesc\footnote{Our current version of this $\mathtt{json}$ is available at: \url{https://github.com/HDI-Project/model-provenance-json}}. 

In our first attempt at creating each of these automation approaches, we focused on defining the right abstractions, process organization and algorithms, because of their foundational role in data science automation. We knew from the beginning the right abstractions would enable us to write code that could be reused across developments, deployments, and domains. This not only increases the rate at which we build models, but increases the likelihood of deployment by enabling the involvement of those with subject matter expertise and those that maintain the production environments. As we have applied our new tools to industrial-scale problems over the last three years, these abstractions and algorithms have transformed into production-ready implementations. With this paper, we are formally releasing multiple tools, as well as data and model organization schemas. They are \ft, \esm, \metajson, \modeldesc, \atm. 

In the sections below, we show how abstractions, algorithms, and implementations all come together to enable ML 2.0.

\section{An ML 2.0 system}

In \mlt, users of the system start with raw untapped data in the warehouse and a rough idea of the problem they want to solve. The user formulates or specifies a prediction problem, follows steps 1-5, and delivers a stable, tested and validated, deployable model in a few weeks, along with a report that documents the \textit{discovery}. 

Although this workflow seems challenging and unprecedented, we are motivated by recent developments in proper abstractions, algorithms, and data science automation tools, which lend themselves to the restructuring of the entire predictive modeling process. In Figure~\ref{fig:new_workflow}, we provide an end-to-end recipe for \mlt. We next describe each step in detail, including its input and output, its automated aspects, and relevant hyperparameters that allow developers to control the process. We highlight both what the human contributes and what the automation enables, illustrating how they form a perfect symbiotic relationship. In an accompanying paper, we present the first industrial use case that follows this paradigm, as well as the accompanying successfully deployed model.


\subsection{Step 1 $\rightarrow$ Form an \es (Data organization)}\label{es}
An \ES is an unified \api for organizing and querying relational datasets to facilitate building predictive models. By providing a single \api, an \es can be reused throughout the life cycle of a modeling project and reduce the possibility of human error while manipulating data. An \es is integrated with the typical industrial-scale data warehouse or data lake that contains multiple tables of information, each with multiple fields or columns. 

The first, and perhaps the most human-driven, step of a modeling project consists of identifying and extracting the subset of tables and columns that are relevant to the prediction problem at hand from the data source. Each table added to the \es is called an entity, and has at least one column, called the \ind, that uniquely identifies each row. Other columns in an entity can be used to relate instances to another entity, similar to a foreign-key relationship in a traditional relational database. An \es further tracks the semantic variable types of the data (e.g the underlying data may be stored as an float, but in context it is important to know that data represents a latitude or longitude).

In predictive modeling, it is common to have at least one entity that changes over time as new information become available. This data is appended as a row to the entity it is an instance of, and must contain a column which stores the value for the time that this particular row became available. This column is known as the \tind for the entity, and is used by automation tools to understand the order in which information became available for modeling. Without this functionality, it would not be possible to correctly simulate historical states of data while training, testing, and validating the predictive model. 

The \ind, \tind, relationships, and semantic variable types for each entity in an \es are documented in \mlt using a \metajson \footnote{Documentation is available at \url{https://github.com/HDI-Project/MetaData.json}}. After extracting and annotating the data, an \es provides the interface to this data for each of the modeling steps ahead.

The open source library \ft offers an in-memory implementation of the \es API in Python to be used by humans and automated tools \footnote{Documentation is available at \url{https://docs.featuretools.com/loading_data/using_entitysets.html}}. \ft represents each entity as a Pandas DataFrame, provides standardized access to metadata about the entities, and offers an existing list of semantic types for variables in the dataset. It offers additional functionality that is pertinent to predictive modeling, including:

\begin{itemize}
    \item[--] querying by time - select a subset of data across all entities up until a point in time (uses time indices)
    \item[--] incorporating new data - append to existing entities as new data becomes available or is collected
    \item[--] checking data consistency - verify new data matches old data to avoid unexpected behavior
    \item[--] transforming the relational structure - create new entities and relationships through normalization (i.e create a new entity with repeated information from an existing entity and add a relationship)
\end{itemize}


\subsection{Step 2 $\rightarrow$ Assemble training examples (Prediction engineering)}
To learn an ML model that can predict a future outcome, past occurrences of that outcome are required in order to train it. In most cases, defining which outcome one is interested in predicting is itself a process riddled with choices. For example:
\begin{quotation}
\textit{Consider a \dev maintaining an online learning platform. He wants to deploy a predictive model predicting when a student is likely to drop out. But dropout could be defined in several ways: as ``student never returns to the website,'' or ``student stops to watch videos,'' or ``student stops submitting assignments.'' }
\end{quotation}

\noindent \textbf{Labeling function}: Our first goal is to provide a way of describing the outcome of interest easily. This is achieved by allowing the \dev to write a \lf function given by 
\begin{equation}\label{outcome}
\mathtt{label, cutoff\_time = f(E, e_{i}, timestamp, hparams)}
\end{equation}

\noindent where, $\mathtt{e_{i}}$ is the $id$ of the $i^{th}$ instance of the entity ($i^{th}$ student, in the example above), $\mathtt{E}$ is \es, and $\mathtt{timestamp}$ is the time point \textit{in the future} at which we are evaluating whether or not the outcome occurred. It is worth emphasizing that this outcome can be a complex function that uses a number of columns in the data. The output of the function is the $\mathtt{label}$, which the predictive model will learn to predict using data up until $\mathtt{cutoff\_time}$.

\noindent \textbf{Search}: Given the \lf function, the next task is to search the data to find past occurrences of the outcome in question. One can simply iterate over multiple instances of the entity at different time points, and call \ref{outcome} in each iteration to generate a label.

\begin{table*}
\centering
\renewcommand{\arraystretch}{1.6}
\begin{tabularx}{\linewidth}{@{}r:X@{}}
\hline
parameter & description \\
\hline
\texttt{prediction\_window} & the period of time to look for the outcome of interest  \\
\texttt{lead} & the amount of time before the prediction window to make a prediction \\
\texttt{gap} & the minimum amount of time between successive training examples for an single instance \\

\texttt{examples\_per\_instance} & maximum number of training examples per entity instance  \\
\texttt{min\_training\_data} & the minimum amount of training required before making a prediction \\
\hline
\end{tabularx}\label{pengparams}
\caption{Hyperparameters for the prediction engineering}
\end{table*}

A number of conditions are applied to this search. Having executed this search for several different problems across many different domains, we were able to precisely and programatically describe this process and expose all of the hyperparameters that a \dev may want to control during this search. Table~\ref{pengparams} presents a subset of these hyperparameters, and we refer an interested reader to \cite{kanter2016label} for a detailed description of these and other search settings.

Given the \lf function, the settings for search, and the \es, a search algorithm searches for past occurrences across all instances of an entity and returns a list of three tuples called label-times:

\begin{equation}
\mathtt{e_{1...n}, label_{1...n}, cutoff\_time_{1...n} = search\_training\_examples(E,f(.), hparams)}
\end{equation}


\noindent where $e_{i}$ is the $i^{th}$ instance of the entity, $\mathtt{label_{i}}$ is a binary 1 or 0 (or multi-category depending upon the \lf) representing whether or not the outcome of interest happened for the $i^{th}$ instance of the entity, and \cof{$_i$} represents the time point after which $\mathtt{label_{i}}$ is known. A list of these tuples provides the possible training examples. While this process is applied in any predictive modeling endeavor, be it in healthcare, educational data mining, or retail, there have been no prior attempts made to abstract this process and provide widely applicable interfaces for it. 

\vspace{1ex}

\noindent
\begin{tabularx}{\linewidth}{@{}>{\itshape\arraybackslash}l@{\hspace{1em}$\to$\hspace{1em}}X@{}}
\toprule
\Sme  & implements the \lf function and sets the hyperparameters. \\
Automation & searches for and compiles a list of training examples that satisfy a number of constraints.\\
\bottomrule
\end{tabularx}


\subsection{Step 3 $\rightarrow$ Generate features (Feature engineering)}
The next step in machine learning model development involves generating features for each training example. Consider:
\begin{quotation}
\textit{At this stage, the \dev has a list of those students who dropped out of a course and those who did not, based on his previously specified definition of dropout. He posits that the amount of time a student spends on the platform could predict whether he is likely to drop out. Using data prior to the \cof, he writes software and computes the value for this feature.}
\end{quotation}

Features, a.k.a variables, are quantities derived by applying mathematical operations to the data associated with an entity-instance -- \textit{but only prior to the \cof as specified in the label\_times}. This step is usually done manually and takes up a lot of \dev time. Our recent work enables us to automate this process by allowing the \dev to specify a set of hyperparameters for an algorithm called Deep Feature Synthesis \cite{kanter2015deep}. The algorithm exploits the relational structure in the data and applies different statistical aggregations at different levels \footnote{An open source implementation of this is available via \url{www.featuretools.com}.} \footnote{A good resource is at: \url{https://www.kdnuggets.com/2018/02/deep-feature-synthesis-automated-feature-engineering.html}}.

\begin{table*}
\centering
\renewcommand{\arraystretch}{1.6}
\begin{tabularx}{\linewidth}{@{}r:X@{}}
\hline
hyperparameter & description \\
\hline 
$\mathtt{target\_entity}$ & entity in \es to create features for  \\
$\mathtt{training\_window}$ & amount of historical data before \cof used to calculate features  \\
$\mathtt{aggregation\_primitives}$ & reusable functions that create new features using data at the intersection of entities\\
$\mathtt{transform\_primitives}$ & reusable functions that create new features from existing features within an entity \\
$\mathtt{ignore\_variables}$ & variables in an \es that shouldn't be used for feature engineering \\
\hline 
\end{tabularx}
\caption{Hyperparameters for feature engineering}
\end{table*}


\begin{equation}\label{fl}
\mathtt{feature\_list = create\_features(E, hparams)}
\end{equation}
\begin{equation}\label{fm}
\begin{split}
& \mathtt{feature\_matrix} = \\
& \mathtt{calculate\_feature\_matrix(E,e_{i\dots n}, cutoff\_time_{i\dots n},label_{i \dots n}, feature\_list, hparams)} 
\end{split}
\end{equation}

\begin{figure*}
\centering
\includegraphics[width=0.95\linewidth]{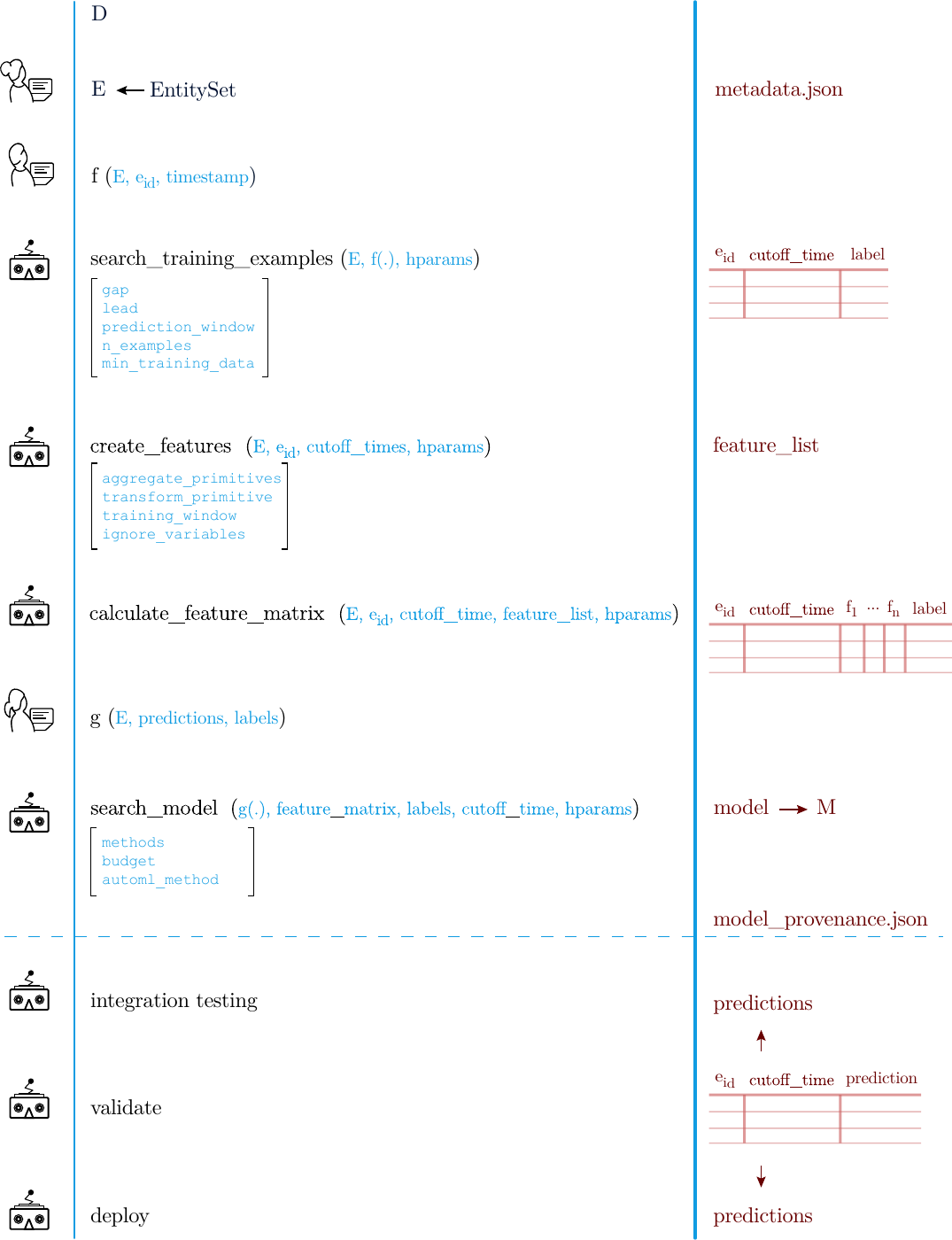}
\caption{ML 2.0 workflow. The end result is a stable, validated, tested deployable model. An ML 2.0 system must support each of these steps. In the middle, we show each of the steps, along with the well-defined, high-level, functional \api{s} and arguments (hyperparameters) that allow for exploration and discovery. Data transformation \api{s} do not change in deployment, ensuring a repeatable process. The rightmost column presents the result of each step.} \label{fig:new_workflow}
\end{figure*}

\noindent Given the \es and hyperparameter settings, the automatic feature generation algorithm outputs a \fl containing a list of feature descriptions defined for the $\mathtt{target\_entity}$ using \linebreak $\mathtt{transform\_primitives}$, $\mathtt{aggregate\_primitives}$, and $\mathtt{ignore\_variables}$. These definitions are passed to \ref{fm} which generates a matrix where each column is a feature and each row pertains to an entity-instance $e_i$ at the corresponding $cutoff\_time_{i}$ and $\mathtt{training\_window}$. The format of the \fm is also shown in Figure~\ref{fig:new_workflow}. The \fl is stored as a serialized file for use in deployment. Users can also apply a feature selection method at this stage to reduce the number of features.

\vspace{1ex}

\noindent
\begin{tabularx}{\linewidth}{@{}>{\itshape\arraybackslash}l@{\hspace{1em}$\to$\hspace{1em}}X@{}}
\toprule
\Sme & guides the process by suggesting which primitives to use or variables to ignore, as well as how much historical data to use to calculate the features.\\ 
Automation & suggests the features based on the relational structure, and  precisely calculates the features for each training example within the allotted window \cof - $\mathtt{training\_window}$ - \cof.\\ 
\bottomrule
\end{tabularx}

\subsection{Step 4 $\rightarrow$ Generate a model $\mathtt{M}$ (Modeling and operationalization)}
Given the $\mathtt{feature\_matrix}$ and its associated labels, the next task is to learn a model. The goal of training is to use examples to learn a model that will generalize well to  future situations, which will be evaluated using a validation set. Typically, a model is not able to predict accurately for all training examples, and must make trade-offs.  While the training algorithm optimizes a \textit{loss} function, metrics like fscore, aucroc, recall, and precision are used for model selection and evaluation on a validation set. These metrics are derived solely from counting how many correct and incorrect \pred are made within each class. In practice, we find that these measures are not enough to capture domain-specific needs, a conclusion we expand upon below:
\begin{quotation}
\textit{Consider a \dev working on a fraud prediction problem. S/he compares two models, $\mathtt{M_1}$ and $\mathtt{M_2}$. While both have the same false positive rate, $\mathtt{M_1}$ has a 3\% higher true positive rate -- that is, it was able to detect more fraudulent activities. When fraud goes undetected, the bank must reimburse the customer -- so the \dev decides to measure the financial impact by adding the cost of the fraudulent transactions that were missed by $\mathtt{M_1}$ and $\mathtt{M_2}$ respectively. He finds that, by this metric, $\mathtt{M_1}$ actually performs worse. Ultimately, the number of frauds $\mathtt{M_2}$ missed does not matter, because missing a fraudulent transaction that is worth \$10 is far less costly than missing a \$10,000 transaction.}
\end{quotation}

This problem can be handled by using a domain specific cost function, $g(.)$, implemented by the \dev.


\noindent \textbf{Cost function $g(.)$}: Given  \pred and true $\mathtt{labels}$ and the \es, a cost function calculates the domain-specific evaluation of the model's performance. Its abstraction is specified as:

\begin{equation}
\mathtt{cost = g(E, predictions,labels)}
\end{equation}


\noindent \textbf{Search for a model}: Given the cost function, we can now search for the best possible model, $M$. An overwhelming number of possibilities exist, and the search space includes all possible methods. In addition, each method contains hyperparameters, and choosing certain hyperparameters may necessitate choosing even more in turn. For example:
\begin{quotation}
\textit{A \dev is ready to finally learn a model $\mathtt{label \gets M(X)}$. He must choose between numerous modeling techniques, including $\mathtt{svm}$, $\mathtt{decision trees}$, and $\mathtt{neural networks}$. If $\mathtt{svm}$ is chosen, s/he must then choose the $\mathtt{kernal}$, which could be $\mathtt{polynomial}$, $\mathtt{rbf}$, or $\mathtt{Gaussian}$. If $\mathtt{polynomial}$ is chosen, then he has to choose another hyperparameter-- the value of the degree.} 
\end{quotation}

\begin{table*}
\centering
\renewcommand{\arraystretch}{1.6}
\begin{tabularx}{\linewidth}{@{}r:X@{}}
\hline
parameter & description \\
\hline 
\texttt{methods} & list of machine learning methods to search  \\
\texttt{budget} & the maximum amount of time or number of models to search \\
\texttt{automl\_method} & path to file describing the automl technique to use for optimization \\
\hline 
\end{tabularx}
\caption{Hyperparameters for model search process}
\end{table*}


 To fully exploit the search space, we designed an approach in which humans fully and formally specify the search space by creating a \texttt{json} specification for each modeling method (see Figure~\ref{method-spec} for an example). The automated algorithm then searches the space for the model that optimizes the $\mathtt{cost}$ assessed using predictions. \cite{swearingenatm}\footnote{Once the modeling method has been chosen, numerous search algorithms for the following steps have been published in academia and/or are available as open source tools.}

\begin{equation}
\mathtt{M = search\_model(g(.), feature\_matrix, labels, cutoff\_times, hparams)}
\end{equation}

To search for a model, based on the cost function $g(.)$, we split the \fm into three sets: a set to train the model, a set to tune the decision function, and a set to test/validate tye model. These splits can be made based on time, and could also be made much earlier in the process, allowing for different label search strategies to extract training examples from different splits. A typical model search works as follows:
\begin{itemize}
    \item[--] Train a model $M_i$ on the training split
    \item[--] Use the trained model to make predictions on the threshold tuning set. 
    \item[--] Pick a decision-making threshold such that it minimizes the cost function $g(.)$
    \item[--] Use the threshold and the model $M$ to make predictions on the test split and evaluate the cost function $g(.)$
    \item[--] Repeat steps 1 - 4 $k$ times for model $M_i$ and determine the average statistics.
    \item[--] Repeat steps 1-5 for different models and pick the one that performs best (on average) on the test split.  
\end{itemize}

The output of this step is a full model specification, including the threshold and the model $M$. $M$ is stored as a serialized file, and the remaining settings are stored in a $json$ called \modeldesc. An example $json$ is shown in Figure~\ref{generic_model_spec_page1}. 

\subsection{Step 5 $\rightarrow$ Integration testing in production}
The biggest advantage of the automation and abstractions designed for the first 4 steps is that they enable an identical and repeatable process with exactly the same \api{s} in the production environment -- a requirement of \mlt. In a production environment, new data is added to the data warehouse $D$ as time goes on. To test a model trained for a specific purpose, after loading the model $M$, the \fl and the \es $E_t$,  a \dev can simply call the following three \api{s}:

\begin{equation}
\mathtt{E_{t+}= add\_new\_data(E_t, new\_data\_path, metadata.json)}
\label{e7}
\end{equation}
\begin{equation}
\mathtt{feature\_matrix= calculate\_feature\_matrix(E_{t+},<e_{i}, }\\
\mathtt{current\_time>, feature\_list)}
\label{e8}
\end{equation}
\begin{equation}
\mathtt{predictions \gets generate\_predictions(M, feature\_matrix)}
\label{e9}
\end{equation}

This allows the developer to test the end-to-end software for new data, ensuring that the software will work in a production environment. 

\subsection{Step 6 $\rightarrow$ Validate in production}
No matter how robustly the model is evaluated during the training process, a robust practice also involves evaluating a model in the production environment. This is paramount for establishing trust, as it identifies issues such as whether the model was trained on a biased data sample, or if the data distributions have shifted since the time the data was extracted for steps 1-4. A \mlt system, after loading the model $M$ and the \fl and undergoing an update to the \es $E_{t+}$, enables this evaluation by a simple tweaking of parameters within the \api{s}:

\begin{align}
\begin{split}
&\mathtt{feature\_matrix\ } \\
&\mathtt{= calculated\_feature\_matrix(E_{t+},<e_{i},arbitrary\_timestamp>,feature\_list)}
\end{split}
\end{align}
\begin{equation}
\mathtt{label, arbitrary\_timestamp = f(E_{t+}, <e_{i}, arbitrary\_timestamp>, hparams)}
\end{equation}
\begin{equation}
\mathtt{predictions \gets generate\_predictions(M, feature\_matrix)}
\end{equation}

\begin{equation}
\mathtt{cost=g(predictions,labels,E_{t+})}
\end{equation}

\subsection{Step 7: Deploy}
After testing and validation, a \dev can deploy the model using the three commands specified by Equations \ref{e7}, \ref{e8}, and \ref{e9}.

\subsection{Meta information}
While going through steps 1 -7 we maintain information about what settings/hyperparameters were set for each stage, as well as the results of the modeling stage, in an attempt to maintain full records of the provenance the process. Our current proposal of this provenance is in \modeldesc and is documented at \url{https://github.com/HDI-Project/model-provenance-json}. This allows us to check for data drift and the unavailability of certain columns, and provides a provenance for what was tried during the training phase and did not work. It is also hierarchical, as it points to three other $\mathtt{jsons}$: one that stores the metadata, a second that stores the search space for each of the methods over which a model search was performed, and a third that stores information about the automl method used. 

\section{Why does this matter?}

\begin{table*}
\centering
\renewcommand{\arraystretch}{1.6}
\begin{tabularx}{\linewidth}{@{}r:X@{}}
\hline
item & description \\
\hline 
$\mathtt{metadata.json}$ & file containing a description of the an \es  \\

$\mathtt{label\_times}$ & the list of label training examples and the point in time prediction will occur \\

$\mathtt{feature\_matrix}$ & table of data with one row per label\_times and one column for each feature \\
$\mathtt{M}$ &  serialized model file returned by model search\\

$\mathtt{feature\_list}$ &  serialized file specifying the $\mathtt{feature\_list}$ returned by feature engineering step\\

$f(.)$ & user-defined function used to create label times \\

$g(.)$ &  cost function used during model search\\

\modeldesc & description of pipelines considered, testing results, and final deployable model \\

$\mathtt{predictions}$ &  the output of the model when passed a \fm \\
\hline
\end{tabularx}
\caption{Different intermediate data and domain-specific code generated during the end-to-end process} \label{diff}
\end{table*}

The proposed \mlt system overcomes the significant bottlenecks and complexities currently hindering the integration of machine learning into products. We frame this as the first proposal to \textit{deliberately} break out of \mlo, and anticipate that as we grow, more robust versions of this paradigm, as well as their implementations, will emerge. Powered by a number of automation tools, we have used the system described above to solve two real industrial use cases -- one from Accenture, and one from BBVA. Below, we highlight the most important contributions of the system described above, and more deeply examine each of its innovations.
\begin{itemize}
    \item[--] \textbf{Standardized representations and process}: In the past, converting complex relational data first into a \textit{learning task} and then into \textit{data representations} had largely been considered an ``art.'' The process lacked scientific form and rigor, precise definitions and abstractions, and -- most importantly --  generalizable framework. It also took about 80\% of the average data scientist's time. The ability to structurally represent this process, and to define the algorithms and intermediate representations that make it up, enabled us to formulate \mlt. If we aim to broadly deliver the promise of machine learning, this process, much like any other software development practice, must be streamlined. 
    \item[--]\textbf{The concept of ``time''}: A few years ago, we participated in a KAGGLE competition using our automatic feature generation algorithm \cite{kanter2015deep}. The algorithm automatically applied mathematical operations to data by finding the relationships within it. The process is typically done manually, and doing it automatically meant a significant boost in data scientists' productivity while maintaining competitive accuracy. When we then applied this algorithm to industrial datasets -- keeping its core ability, generating features from data, intact -- we recognized that, in this new context, it was necessary to first define a concrete machine learning task. When a task is defined, performing step 2 (assembling training examples) implied that \textit{``valid''} data had been identified, and since task definition required flexibility, which data was valid and which was not changed each time we changed the definition -- a key problem we identified in \mlo. (Competition websites perform this step before they even release the data.) This inspired us to consider the third and most important dimension of our automation: ``time.'' To identify valid data, we needed our algorithms to filter data based on the time of their occurrence, which meant giving every data point an annotation pinpointing when it was first ``known.''\footnote{This is not necessarily equivalent to the time it was recorded and stored in the database, although sometimes it is.} A complex interplay between the storage and usage of these annotations ensued, eventually resulting in \mlt. Just as our ability to automate the search process led to step 2, accounting for time allows unprecedented flexibility in defining machine learning tasks by maintaining provenance. 
    \item[--]\textbf{Automation of processes}: While our work in early 2013 focused on the automation of model search (step 4), we moved to automating step 3 in 2015, and subsequently step 2 in 2016. These automation systems enabled us to go through the entire machine learning process end-to-end. Once we had fully automated these steps, we were then able to design interfaces and \api{s} that enabled easy human interaction and a seamless transition to deployment.
    \item[--]\textbf{Same \api{s} in training and production}: The \api{s} used in steps 5-7 are exactly the same as the ones used in steps 2 and 4. This implies that the same software stack is used both in training and deployment. This software stack allows the same team to execute all the steps, requires a minimal learning curve when things need to be changed, and most importantly, provides proof of the model's provenance -- how it was validated both in the development environment and in the testing/production environment.
    \item[--]\textbf{No redundant copies of data}: Throughout \mlt, new data representations made up of computations over the data -- including features, training examples, metadata, and model descriptions -- are created and stored. However, nowhere in the process is a redundant copy of a data element made. \footnote{Unless the computation is as simple as identity -- for example, if the age column is available in the data, and we would use this column as a feature as-is, it will appear in the original data as well as in the feature matrix.} Arguably, this mitigates a number of data management and provenance issues elucidated by researchers. Although it is hard to say whether all such issues will be mitigated across the board, \mlt eliminates multiple hangups that are usually a part of machine learning or predictive modeling practice. 
    \item[--]  \textbf{Perfect interplay between humans and automation}: After deliberate attempts to take our tools and systems to industry, and debates over \textit{who it is we are enabling} when we talk about "democratize machine learning," we have come to the conclusion that a successful \mlt system will require perfect interplay between humans and  automated systems. Such an interplay ensures that: (1) Humans control aspects that are closely related to the domain in question -- such as domain-specific cost functions, the outcome they want to predict, how far in advance they want to predict, etc. -- which are exposed as hyperparameters set in steps 2-5. (2) The process abstracts away the nitty-gritty details of maching learning, including data processing based on the various forms and shapes in which it is organized, generating data representations, searching for models, provenance, testing, validation and evaluation. (3) A structure and common language for the end-to-end process of machine learning aids in sharing 
    
\end{itemize}

\section{The first AI product - delivered using \mlt}
In an accompanying paper called ``The AI Project Manager'', we demonstrate how we used the seven steps above to deliver a new AI product for the global enterprise Accenture. Not only did this project take up an ML problem that didn't have an existing solution, it presented us an unique opportunity to work with \sme{s} to develop a system that would be put to use. In this case, the goal was to create an AI Product Manager that could augment human product managers while they manage complex software projects. Using over 3 years of historical data, which contained 40 million reports and 3 million comments, we trained a machine learning-based model to predict, weeks in advance, the performance of software projects in terms of a host of delivery metrics. The \sme{s} provided these delivery metrics for prediction engineering, as well as valuable insights into the subsets of data that should considered for automated feature engineering. Under the ML 2.0 process, the end-to-end project spanned 8 weeks. In live testing, the AI Project Manager correctly predicts potential issues 80\% of the time, helping to improve key performance indicators related to project delivery. As a result, the AI Project Manager has been integrated in Accenture’s myWizard Automation Platform and serves predictions on a weekly basis.



 %



\section{Commentary on ML 1.0}\label{commentary}

Perhaps the fundamental problem with the ML 1.0 workflow is that it was never intended to create a usable model. Instead, it was designed to make \textit{discoveries}. In order to achieve the end goal of an analytic report or research paper introducing the \textit{discovery}, numerous usability tradeoffs are made. Below, we highlight some of these tradeoffs, and how they are at odds with creating a deployable, useful model. 

\begin{itemize}
    \item[--] \textbf{Ad hoc, non-repeatable software engineering practice for steps 1- 4}: The software that carries out steps 1 - 4 is often ad-hoc and lacks a robust, well-defined software stack. This is predominantly due to a lack of recognized structures, abstractions and methodologies. Most academic research in machine learning has ignored this lack for the past 2 decades, and only focused on defining a robust practice when the data is ready to be used at step 5, and/or the data fits the form of a clean, time-stamped time series $X,t$ (for example, the time series of a stock price). \footnote{Take for example, ``Elements of Statistical Learning,'' a well-recognized and immensely useful textbook that describes numerous methods and concepts relevant only to data in a matrix form.} This implies that a subset of transformations done to the data during \dis will have to be re-implemented later -- probably by a different team, requiring extensive back-and-forth communication.  
    
    \item[--] \textbf{Ad hoc data practices}: In almost all cases, as users go through steps 1, 2, and 3 and insert their domain knowledge, the data output at each stage is stored on a disk and passed around with some notes (sometimes through emails). To ensure confidence in discoveries made, data fields that are questionable or cannot be explained easily or without caveats are dropped. No provenance is maintained for the entire process. 
    
    \item[--] \textbf{Siloed and non-iterative}: These steps are usually executed in a siloed fashion, often by different individuals within an enterprise. Step 1 is usually executed by people managing the data warehouse. Step 2 is done by domain experts in collaboration with the software engineers maintaining the data warehouse. Steps 3 and 4 fall under the purview of data science and machine learning experts. At each step, a subset of intermediate data is stored and passed on. Once a step is finished, very rarely does anyone wants to revisit it or ask for additional data, as this often entails some kind of discussion or back-and-forth.
    
    \item[--] \textbf{Over-optimizing steps 4 and 5, especially step 5}: While the data inputs for steps 1 and 2 are complicated and nuanced, the data is usually simplified and shrunk down to a goal-specific subset by the time it is arrives at step 4. It is simplified further when it arrives at step 5, at which stage it is just a matrix. Steps 1, 2 and 3 are generally considered tedious and complex, and so when work is done to bring the data to step 4, engineers make the most of it. This has led to steps 4 and 5 being overemphasized, engineered and optimized. For example, all data science competitions, including those on \textsc{Kaggle}, start at Step 4, and almost all machine learning startups, tools and systems start at Step 5. At this point, there are several ways to handle these two steps, including crowdsourcing, competitions, automation, and scaling. Sometimes, these over-engineered solutions do not translate to real-time use environments. Take the 2009 Netflix challenge: The winning solution was comprised of a mixture of >100 models, which made it difficult to engineer for real-time use \footnote{https://medium.com/netflix-techblog/netflix-recommendations-beyond-the-5-stars-part-1-55838468f429}. 
    \item[--] \textbf{Departure from the actual system collecting the data}: Since data is pulled out from the warehouse at time $t$, and the \textit{discovery} ensues at some arbitrary time in the future, it may be the case that the system is no longer recording certain fields at that time. This could be due to policy shifts, the complexity or cost involved in making those recordings, or simply a change in the underlying product or its instrumentation. While everything written as part of the \textit{discovery} is still valid, the model may not actually be completely implementable, and so might not be put to use. 
\end{itemize}


Ultimately, a \textit{discovery} has to make its way to real use by integrating with real-time data. This implies that new incoming data has to be processed/transformed in the precise way it was to learn the model, and that a robust prediction-making software stack must be developed and integrated with the product. Usually, this responsibility falls on the shoulders of the team responsible for the product, who most likely have not been involved in the end-to-end model discovery process. Thus the team starts auditing the discovery, trying to understand the process while simultaneously building the software, learning concepts that may not be in their skill set (for example, cross validation), and attempting to gain confidence in the software as well as its predictive capability, all before putting it to use. Lacking of well-documented steps, robust practice for steps 1-4, and the ability to replicate a model accuracy can often result in confusion, distrust in the possibility of the model delivering the intended result and ultimately abandonment of deployment altogether. 



\newpage

\section*{Appendix}
\captionof{figure}{\label{method-spec}A specification of search parameters for a machine learning method.}
\begin{lstlisting}[language=JavaScript]
{   
    "name": "dt",
    "class": "sklearn.tree.DecisionTreeClassifier",
    "parameters": {
        "criterion": {
            "type": "string",
            "range": ["entropy", "gini"]
        },
        "max_features": {
            "type": "float",
            "range": [0.1, 1.0]
        },
        "max_depth": {
           "type": "int",
           "range": [2, 10]
        },
        "min_samples_split": {
           "type": "int",
           "range": [2, 4]
        },
        "min_samples_leaf": {
           "type": "int",
           "range": [1, 3]
        }
    },
    "root_parameters": ["criterion", "max_features", 
    "max_depth", "min_samples_split", "min_samples_leaf"],
    "conditions": {}
}
\end{lstlisting}

\captionof{figure}{\label{generic_model_spec_page1}
An example JSON prediction log file detailing the provenance of the accompanying model, including instructions for labeling, data splitting, modeling, tuning, and testing (continued in Figure~\ref{generic_model_spec_page2}). The most up-to-date specification of this json can be found at https://github.com/HDI-Project/model-provenance-json.}
\begin{lstlisting}[language=JavaScript]
{
    "metadata": "/path/to/metadata.json",

    "prediction_engineering": {
        "labeling_function": "/path/to/labeling_function.py",
        "prediction_window": "56 days",
        "min_training_data": "28 days",
        "lead": "28 days"
    },

    "feature_engineering":[
      {
        "method": "Deep Feature Synthesis",
        "training_window": "2 years",
        "aggregate_primitives": ["TREND", "MEAN", "STD"],
        "transform_primitives": ["WEEKEND", "PERCENTILE"],
        "ignore_variables":{
            "customers": ["age", "zipcode"],
            "products": ["brand"]
        },
        "feature_selection": {
            "method": "Random Forest",
            "n_features": 20
        }
      }
    ],

    "modeling": {
        "methods": [{"method": "RandomForestClassifer",
                     "hyperparameter_options": "/path/to/random_forest.json"},
                    {"method": "MLPClassifer",
                     "hyperparameter_options": "/path/to/multi_layer_perceptron.json"}],
        "budget": "2 hours",
        "automl_method": "/path/to/automl_specs.json",
        "cost_function": "/path/to/cost_function.py"
    },
    
    "data_splits": [
        {
            "id": "train",
            "start_time": "2014/01/01",
            "end_time": "2014/06/01",
            "label_search_parameters": {
                "strategy": "random",
                "examples_per_instance": 10,
                "offset": "7 days",
                "gap": "14 days"
            }
        },
        {
            "id": "threshold-tuning",
            "start_time": "2014/06/02",
            "end_time": "2015/01/01",
            "label_search_parameters": {"offset": "7 days"}
        },
        {
            "id": "test",
            "start_time": "2015/01/02",
            "end_time": "2015/06/01",
            "label_search_parameters": {"offset": "7 days"}
        }
    ],
\end{lstlisting}

\captionof{figure}{\label{generic_model_spec_page2}An example JSON prediction log file detailing the provenance of the accompanying model, including instructions for labeling, data splitting, modeling, tuning, and testing (continued from Figure~\ref{generic_model_spec_page1})}
\begin{lstlisting}[language=JavaScript]
    "training_setup": {
        "training": {"data_split_id": "train",
                     "validation_method": "/path/to/validation_spec_train.json"},
        "tuning": {"data_split_id": "threshold-tuning",
                   "validation_method": "/path/to/validation_spec_tune.json"},
        "testing": {"data_split_id": "test",
                    "validation_method": "/path/to/validation_spec_test.json"}
    },


    "results": {
      "test": [
          {
              "random_seed": 0,
              "threshold": 0.210,
              "precision": 0.671,
              "recall": 0.918,
              "fpr": 0.102,
              "auc": 0.890
          },
          {
              "random_seed": 1,
              "threshold": 0.214,
              "precision": 0.702,
              "recall": 0.904,
              "fpr": 0.113,
              "auc": 0.892
          }
      ]
    },


    "deployment": {
        "deployment_executable": "/path/to/executable",
        "deployment_parameters": {
            "feature_list_path": "/path/to/serialized_feature_list.p",
            "model_path": "/path/to/serialized_fitted_model.p",
            "threshold": 0.212
        },
        "integration_and_validation": {
            "data_fields_used": {
                    "customers": ["name"],
                    "orders": ["Order Id", "Timestamp"],
                    "products": ["Product ID", "Category"],
                    "orders_products": ["Product Id", "Order Id", "Price", "Discount"]
            },
            "expected_feature_value_ranges":{
                "MEAN(orders_products.Price)": {"min": 9.50, "max": 332.30},
                "PERCENT(WEEKEND(orders.Timestamp))": {"min": 0, "max": 1.0}
            }
        }
}
\end{lstlisting}

\begingroup
\parindent 0pt
\parskip 2ex
\def\enotesize{\normalsize}
\endgroup
%

 
\printglossaries

\end{document}